\documentstyle{article}
\begin{document}

\title{{\small Appears in the Working Notes of the AAAI Spring Symposium on
Representation and Acquisition of Lexical Knowledge: Polysemy,
Ambiguity, and Generativity, Palo Alto, CA, March 27--29, 1995\\}
\bf Lexical Acquisition via Constraint Solving
	\thanks{Supported in part by the National Science Foundation
                under Grant No.~IRI-9212074.}}

\author{Ted Pedersen \ \ \  Weidong Chen\\
        \small Department of Computer Science \& Engineering\\
        \small Southern Methodist University\\
        \small Dallas, TX 75275\\
        \small\tt \{pedersen,wchen\}@seas.smu.edu}

\date{}

\maketitle

\begin{abstract}
This paper describes a method to automatically acquire the syntactic
and semantic classifications of unknown words. Our method reduces the
search space of the lexical acquisition  problem by
utilizing both the left and the right context of the unknown
word. Link Grammar provides a convenient framework in which to
implement our method.
\end{abstract}

\section{Introduction}

A robust Natural Language Processing (NLP) system must be able to
process sentences that contain words unknown to its lexicon. The
syntactic and semantic properties of unknown words are derived from
those of known words in a sentence, assuming that the given sentence
is valid.

The underlying linguistic framework plays a critical role in lexical
acquisition. Linguistic frameworks can be broadly classified into two
groups: those with phrase structure rules and those without.
The lexicon of known words and any phrase structure rules that exist
determine the size of the search space for the classification of
unknown words. In general, the more complex the phrase structure
rules, the larger the search space.

This paper explores lexical acquisition in a framework without phrase
structure rules. All constraints on the usage of words are integrated
into the lexicon. We use a novel lexical representation that
explicitly specifies what syntactic and semantic classes of words may
appear to the left and to the right of a word in a valid sentence.
If all words are known in a sentence, it is valid only if
the associated constraints have a solution. Otherwise, constraints are
inferred for unknown words that will make the sentence valid.

We choose to use Link Grammar\cite{SleatorT91} as it
provides a convenient means for
expressing bidirectional constraints. Among the other frameworks we have
investigated were Dependency Grammar \cite{Melcuk88}, Categorial
Grammar \cite{OehrleBW88}, and Word Grammar \cite{Hudson84} all of which
are lexically--based. We selected Link Grammar due to its explicit use
of right and left context and the availability of an implementation that
includes a 24,000 word lexicon. However, our approach is
applicable to any system that integrates bidirectional constraints
explicitly in the lexicon.

This paper begins with an introduction to Link Grammar. We describe the
process of acquiring the syntax of unknown words and outline the
process of semantic acquisition. We close with a discussion of related
work and our plans for the future.

\section{Link Grammar}

Link Grammar\cite{SleatorT91} is a context--free linguistic framework
that is lexically based. It differs from other context--free grammars
in that there are no decomposable constituent structures and its
grammar rules are implicit in the lexicon.

Each word in the grammar is defined by a syntactic constraint that is
expressed in a disjunctive normal form. Each disjunct consists of a pair of
ordered lists of the form
$((l_1,...,l_{m-1},l_m)(r_n,r_{n-1},...,r_1))$
where the left hand list is made up of connectors that must link to
words to the left of the word in the sentence  and likewise for the
right hand list. Each word can have multiple disjuncts, which implies
that it can be used in various syntactic contexts.

The following is a simple example of a Link Grammar:

\begin{quote}
\begin{tabbing}
WWWWWWWWW  \= Link  \kill

big, yellow: \> (( ) (A)) \\

car, corn, condor, \\
gasoline, meat: \>   ((A,Ds,Os) ( )) \\
                     \>   ((A,Ds) (Ss))   \\
                     \>   ((Ds) (Ss))    \\
                     \>   ((Ds,Os) ( ))    \\
                     \>   ((Os) ( ))    \\
eats: \>  ((Ss) (O)) \\
      \>  ((Ss) ( )) \\
the:  \>  (( ) (D)) \\
\end{tabbing}
\end{quote}

Parsing a sentence in Link Grammar consists of choosing one disjunct
for each word such that it can be  connected to the surrounding words
as specified in that disjunct.
For a simple example consider the sequence of words: ``The condor eats
the meat''
and the following  choices of disjuncts for each word from the lexicon above:
\begin{quote}
\begin{tabbing}
WWWWWWWWW  \= Link  \kill
the: \>    (( ) (D)) \\
condor: \>  ((Ds) (Ss)) \\
eats: \>  ((Ss) (O)) \\
the:  \>   (( ) (D)) \\
meat: \>   ((Ds,Os) ( )) \\
\end{tabbing}
\end{quote}
The following diagram (called a linkage) shows the
links among the words that justify the validity of the sentence
according to Link Grammar.

\begin{verbatim}
                     +----Os--+
         +-Ds--+-Ss--+   +-Ds-+
         |     |     |   |    |
        the condor eats the meat
\end{verbatim}

In general, a sequence of words is a sentence if it is possible to
draw links among the words in such a way that the syntactic constraint
of every word is satisfied and all the following meta--rules are
observed:

\begin{itemize}
\item {Planarity:} Links drawn above the sentence do not intersect.

\item {Connectivity:} There is a path from any word in the sentence to
any other word via the links.

\item {Ordering:} For each disjunct of a word $w$, of the form
	$((l_1,...,l_{m-1},l_m)(r_n,r_{n-1},...,r_1))$, where
	$m \ge 0$ and $n \ge 0$, the left hand list of connectors
	indicates links to words to the left of $w$, and likewise for the
	right hand list. In addition,
	the larger the subscript of a connector, the further
	away the word with the matching connector is from $w$.

\item {Exclusion:} No two links may connect the same pair of words.
\end{itemize}
Parsing in Link Grammar corresponds to constraint solving according to
these meta--rules. The objective is to select one disjunct for each
word in a sentence that will lead to satisfaction of the the meta--rules.

\section{Syntactic Acquisition}

Syntactic acquisition is the process of mapping an unknown  word to a
finite
set of syntactic categories. In Link Grammar syntactic categories are
represented by the constraints that are expressed as disjuncts.
Our lexical acquisition system is not called upon
to create or identify new syntactic categories as we assume that these
are already known.

Given a sentence with unknown words the disjuncts
of unknown words are determined based upon
the syntactic constraints of the known words in the sentence.

For instance suppose that {\em snipe} is an unknown word in the
sentence: ``The {\em snipe\/} eats meat''. The following lists all the
choices for the disjuncts of the known words which come from the lexicon.

\begin{quote}
\begin{tabbing}
WWWWWWWWW  \= Link  \kill
the:  \>  (( ) (D)) \\
snipe:  \>  ((?) (?)) \\
eats: \>  ((Ss) (O)) \\
      \>  ((Ss) ( )) \\
meat: \>   ((A,Ds,Os) ( )) \\
      \>   ((A,Ds) (Ss))   \\
      \>   ((Ds) (Ss))    \\
      \>   ((Ds,Os) ( ))    \\
      \>   ((Os) ( ))
\end{tabbing}
\end{quote}

It must be determined what disjunct associated with `snipe' will allow for
the selection of a single disjunct for every known word such that each
word can have its disjunct satisfied in accordance with the
meta--rules previously discussed. There are 10 distinct disjuncts in
the above grammar and any one of those could be the proper syntactic
category for `snipe'.

We could attempt to parse by blindly assigning to `snipe' each of
these disjuncts and see which led to a valid linkage. However this is
impractical since more complicated grammars will have hundreds or even
thousands of known disjuncts. In fact, in the current 24,000 word lexicon there
are
approximately 6,500 different syntactic constraints. A
blind approach would assign all of these disjuncts to `snipe' and then
attempt to parse.
It is possible to greatly reduce the number of candidate disjuncts by
analyzing the disjuncts for the known words.
Those disjuncts that violate the constraints of the meta-rules are
eliminated.

The disjuncts ((A,Ds)(Ss)) and ((Ds)(Ss))
for `meat' are immediately
eliminated as they can never be satisfied since there are
no words to the right of `meat'.

The disjunct ((A,Ds,Os)( )) for `meat'
can also be eliminated. If the A connector is to be
satisfied it would have to be satisfied by `snipe'. The ordering meta--rule
implies that the Ds connector in `meat' would have to be satisfied
by `the' but then the remaining Os connector in
`meat' would not be satisfiable since there are no words preceding
`the'.

That leaves the disjuncts ((Ds,Os)( )) and ((Os)( )) as the remaining
possibilities for `meat'. The disjunct ((Ds,Os)( )) can be eliminated
since the only words that can satisfy the Ds connector are `the' or
`snipe'. Again the ordering meta--rule makes it impossible to satisfy the Os
connector. Thus the only remaining candidate disjunct for `meat' is ((Os)( )).

The next word considered is `eats'. There are two possible disjuncts
and neither can be immediately eliminated. The left hand side of each
disjunct consists of an Ss connector. This could only be satisfied by
`snipe' which therefore must have an Ss connector in its right hand
side. Recall that the left hand side of `meat' consists of an Os
connector. This could be satisfied either by the ((Ss)(O)) disjunct for
`eats' or if the right hand side of `snipe' consists of ((Os,Ss)). The
left hand side of 'snipe' need only consist of a D connector in order
to satisfy the right hand side of `the'. Thus the disjunct for `snipe'
must be either ((D)(Ss)) or ((D)(Os,Ss)) and we have still not eliminated
any of the candidate disjuncts for `eats'.
Unfortunately the meta--rules do
not allow for the further elimination of candidate disjuncts.

In cases such as this the lexicon is used as a knowledge source and
will be used to resolve the issue. The disjunct ((D)(Ss)) is selected
for the word `snipe' since it appears in the lexicon and is normally
associated with simple nouns. Thus the disjunct
((Ss)(O)) is the only possibility for `eats'.

The disjunct ((D)(Os,Ss)) does not appear in the lexicon and
in fact implies that the word it is associated with is both a noun and
a verb.
To eliminate such nonsensical combinations of connectors the
lexicon of known words is consulted to see if a theorized disjunct has
been used with a known word, and if so it is accepted.
The intuition is that even though a word is unknown it is likely to
belong to the same syntactic category as that of some known
words.  This follows from the assumption
that the set of syntactic categories is closed and will not be
added to by the lexical acquisition system. For efficiency these
constraints can be used to
avoid the generation of nonsensical disjuncts in the first place.

To summarize, the following assignment of disjuncts satisfies the
meta--rules and leads to the linkage shown below.

\begin{quote}
\begin{tabbing}
WWWWWWWWW  \= Link  \kill
the:  \>  (( ) (D)) \\
snipe:\>  ((D) (Ss)) \\
eats: \>  ((Ss) (O)) \\
meat: \>   ((Os) ( ))
\end{tabbing}
\end{quote}

\begin{verbatim}
         +-Ds--+-Ss--+--Os-+
         |     |     |     |
        the  snipe eats  meat
\end{verbatim}

\section{Semantic Acquisition}

Acquisition of lexical semantics is defined in
\cite{Berwick83,Granger77,Hastings94,Russell93} as mapping unknown
words to known concepts. \cite{Hastings94,Russell93} assume that the
knowledge base is a concept hierarchy structured as a tree where
children are more specific concepts than their parents.  There are
separate hierarchies for nouns and verbs.
Rather than using concept
hierarchies  \cite{Berwick83,Granger77} used scripts and causal
networks to represent a sequence of related events.
In their work Lexical Acquisition consists
of mapping an unknown word into a known sequence of events.
We adopt the convention of \cite{Hastings94,Russell93} and attempt to
map unknown words into a concept hierarchy.

In order to semantically classify an unknown word the lexical entries
of known words must be augmented with semantic information derived from
the actual usage of them in a variety of contexts.

As sentences with no unknown words are parsed, each connector in the
syntactic constraints of nouns and verbs is tagged with the noun or
verb with which it connects to. For instance given the
sentence: ``The condor eats meat'', the nouns and verbs
are tagged as follows:

\begin{quote}
\begin{tabbing}
WWWWWWWWW  \= Link  \kill
the:  \>  (( ) (D)) \\
condor:  \>  ((D) ($Ss_{eats}$)) \\
eats: \>  (($Ss_{condor}$) ($O_{meat}$)) \\
meat: \>   (($Os_{eats}$) ( ))
\end{tabbing}
\end{quote}

When a word occurs in related syntactic contexts the semantic tags on
the syntactic constraints are merged through generalization using the
superclass information contained in the lexicon. Suppose that the
following sentences with no unknown words have been processed.

\begin{verse}
S1: The big cow eats yellow corn. \\
S2: The condor eats meat. \\
S3: The car eats gasoline. \\
\end{verse}
The corresponding semantic tags for `eats' are:
\begin{quote}
\begin{tabbing}
WWWWWWWWW  \= Link  \kill
eats: \>  (($Ss_{cow}$) ($O_{corn}$)) \\
 \>  (($Ss_{condor}$) ($O_{meat}$)) \\
 \>  (($Ss_{car}$) ($O_{gasoline}$))
\end{tabbing}
\end{quote}

{}From sentences S1 and S2 a more general semantic
constraint is learned since `animal' subsumes `cow' and `condor' and
`food' subsumes `corn' and `meat'. This knowledge is expressed by:

\begin{quote}
\begin{tabbing}
WWWWWWWWW  \= Link  \kill
R1: \>  (($Ss_{animal}$) ($O_{food}$)) \\
 \>  (($Ss_{car}$) ($O_{gasoline}$))
\end{tabbing}
\end{quote}

The semantic tags applied to the connectors serve as semantic
constraints on the words that `eats' connects to. The first disjunct
in the above entry tells us that `eats' must have a concept that is
subsumed by `animal' to its left and a concept that is subsumed by
`food' to its right.

While the lexicon has no information about the unknown word it
does have the semantic constraints of the known words in the
sentence. These are used to infer what the semantic classification of
the unknown word should be if the sentence is valid.

No semantic information has been acquired for `snipe'. If the
nouns and verbs in the sentence, ``The snipe eats meat'',  are tagged
with the nouns and verbs that they connect to, the following is obtained:.

\begin{quote}
\begin{tabbing}
WWWWWWWWW  \= Link  \kill
the:  \>  (( ) (D)) \\

snipe:  \>  ((D) ($Ss_{eats}$)) \\

eats: \>  (($Ss_{snipe}$) ($O_{meat}$)) \\

meat: \>   (($Os_{eats}$) ( ))
\end{tabbing}
\end{quote}

The lexicon has no knowledge of `snipe' but it does have knowledge of
the verb, 'eats', that links to `snipe'. It must be determined which
of the two usages of `eats' described in R1) applies to the
usage of `eats' in ``The snipe eats meat''.

According to the concept hierarchy `meat' is
subsumed by `food' whereas `gasoline' is not. This indicates that the usage
(($Ss_{animal}$)($O_{food}$)) is more appropriate and that `snipe' must
therefore be tentatively classified as an animal. This classification
can be refined as other usages of `snipe' are encountered.

\section{Discussion}

There has been extensive work on lexical acquisition.
Probabilistic part--of-speech taggers have been successful in
identifying the part--of--speech of unknown
words\cite{Church88,WeischedelMSRP93}. These approaches often
require large amounts of manually tagged text to use as training data.

Unknown words have been semantically classified using
knowledge intensive methods\cite{Berwick83,Granger77,Zernik87A}. They
assume the availability of scripts or other forms of detailed
domain knowledge that must be manually constructed. While they have
considerable success in specific domains it is difficult to port such
systems to new domains without requiring extensive manual
customization for the new domain.

Explanation Based Learning has been applied to lexical
acquisition\cite{AskerGS92}. A large corpus of text is divided into
sentences with unknown words and those without. Those without are
parsed and their parse trees form a knowledge base. When a
sentence with an unknown word is processed the system locates the
parse tree that most closely resembles it  and attempts to infer the
syntax of unknown words from this tree. This approach assumes that the
sentences with known words produce parse trees that will match or
cover all of the sentences with unknown words. A limiting factor of
this method is the potentially large number of distinct parse trees.

Unification--based grammars have been brought to bear on the
problem of unknown words\cite{Hastings94,Russell93}. These approaches
are similar in that properties of unknown words are inferred from the
lexicon and phrase structure rules. However, as the underlying
parsers work from left--to--right it is natural to propagate
information from known words to unknown words in the same direction.

The distinctive features to our approach are that all the required
knowledge is represented explicitly in the lexicon and constraint
solving is bidirectional. This makes maximal use of the constraints of
known words and reduces the search space for determining the
properties of unknown words. Link Grammar is not the only way to
process grammars bidirectionally. In fact, there is no reason why a
more traditional context free grammar could not be processed
bidirectionally\cite{SattaS94}.

An implementation is under way to extend the parser of Link Grammar to
automatically acquire the syntax and semantics of unknown words.
It seems that the disjuncts of each word are a special kind of feature
structure. An interesting topic is to integrate feature structures and
unification with Link Grammar to allow more expressive handling of
semantic information.

\end{document}